# A Logical Characterization of Constraint-Based Causal Discovery

**Tom Claassen and Tom Heskes**
Institute for Computer and Information Science
Radboud University Nijmegen
The Netherlands

## Abstract

We present a novel approach to constraint-based causal discovery, that takes the form of straightforward logical inference, applied to a list of simple, logical statements about causal relations that are derived directly from observed (in)dependencies. It is both sound and complete, in the sense that all invariant features of the corresponding partial ancestral graph (PAG) are identified, even in the presence of latent variables and selection bias. The approach shows that every identifiable causal relation corresponds to one of just two fundamental forms. More importantly, as the basic building blocks of the method do not rely on the detailed (graphical) structure of the corresponding PAG, it opens up a range of new opportunities, including more robust inference, detailed accountability, and application to large models.

## 1 Introduction

Causal discovery remains at the heart of most scientic research to date. Understanding which variables in a causal system influence which other is crucial for predicting the effects of actions and policies. Learning such relations from observational data is challenging, especially when latent confounders (hidden common causes) and selection bias (affecting the chance of inclusion in the data set) can be present.

With the introduction of the FCI algorithm in the seminal work of (Spirtes et al., 2000), it was shown that, under reasonable assumptions, it is indeed possible to infer valid causal information from observed probabilistic independencies in the large sample limit. Subsequent results and contributions from various researchers (Spirtes et al., 1999; Ali et al., 2005; Zhang, 2008a) have developed this into a method that produces a provably sound and complete output model that captures all identifiable causal information.

Perhaps surprisingly, this does not mean that the problem of causal discovery from data is now considered 'solved' by the wider research community: the method has trouble handling large models, and worse, in practice the output is often seen as unreliable.

Most current constraint-based approaches to causal discovery rely on a two step process: a structure identification phase from observed independencies, followed by a (graphical) orientation phase. In real-world data, the large sample limit does not apply, and if one or more incorrect independence decisions are made, then this can lead to a series of erroneous orientations. But this ambiguity is not apparent in the output, severely limiting the interpretability of the entire model, also because there is little or no accountability prospect for any of the causal relations found. Bayesian score-based methods such as GES (Chickering, 2002) are better suited to deal with this kind of problem, as they can produce multiple output models (Heckerman et al., 1999). However, they have trouble accounting for hidden variables and selection bias, and also suffer from the complexity problem. Would it be possible to give a (complete) characterization of identifiable causal relations, without first building a global structure?

**Related work**

One of the first to identify causal relations without recourse to a global structure was Cooper (1997), who presented an algorithm that could infer a causal relation from certain independence relations between three variables, in combination with information that one of these was known to be an uncaused variable. Later Mani et al. (2006) showed that certain independence relations between four variables, corresponding to a so-called embedded Y-structure, indicate the presence of a causal relation without the need for such background knowledge. A different approach to tackle the large scale complexity was taken by (Spirtes, 2001), who introduced a variant of FCI that could be interrupted at various stages in the inference process, with an output that is correct, but perhaps less informative than if the algorithm had been allowed to complete.

The perceived lack of robustness in the graphical orientation phase (due to incorrect independence decisions from limited available data), was addressed by Ramsey et al. (2006) who introduced checks to identify certain inconsistencies in the observed independencies that violate so-called 'orientation faithfulness', and avoid propagating these to the rest of the graph. In a related paper, Zhang and Spirtes (2008) also consider testable instances of adjacency unfaithfulness.

In this paper we introduce three rules to convert observed minimal (in)dependencies into logical statements about causal relations. We show that straightforward inference on these logical statements, using standard properties of causality, is sufficient to obtain all identifiable causal information. The result is the first provably sound and complete alternative to the augmented FCI algorithm. As such it generalizes other methods that do not need to build a global independence structure first, and is easily formulated as an anytime algorithm. The fact that the method only requires three simple rules that can be combined in any desired order of occurrence offers hope that this approach can be adapted to improve robustness on real-world data sets as well.

The paper is organized as follows. Section 2 describes some standard methods and terminology. Section 3 introduces the logical inference rules. Sections 4 and 5 show these rules are complete. Proofs are provided in the Appendix and in (Claassen and Heskes, 2011).

## 2 Background

### 2.1 Mixed graphical models

A *mixed graph* $\mathcal{G}$ is a graphical model that can contain three types of edges between pairs of nodes: directed ($\longrightarrow$), bi-directed ($\longleftrightarrow$), and undirected ($\longrightarrow$). In a mixed graph, standard graph-theoretical notions, e.g. *child/parent*, *ancestor/descendant*, *directed path*, *cycle*, still apply. So, a vertex $Z$ is a *collider* on a path $u = \langle \ldots, X, Z, Y, \ldots \rangle$ if there are arrowheads at $Z$ on both edges from $X$ and $Y$, otherwise it is a *noncollider*; if $X$ and $Y$ are not adjacent in $\mathcal{G}$, then the subpath $\langle X, Z, Y \rangle$ is *unshielded*.

A mixed graph $\mathcal{G}$ is *ancestral*, iff an arrowhead at $X$ on an edge to $Y$ implies there is no directed path from $X$ to $Y$ in $\mathcal{G}$, and there are no arrowheads at nodes with undirected edges. As a result, arrowhead marks can be read as 'is not an ancestor of'. In a mixed graph $\mathcal{G}$, a vertex $X$ is *m-connected* to $Y$ by a path $u$, relative to a set of vertices $\mathbf{Z}$, iff every noncollider on $u$ is not in $\mathbf{Z}$, and every collider on $u$ is an ancestor of $\mathbf{Z}$. If there is no such path, then $X$ and $Y$ are *m-separated* by $\mathbf{Z}$. An ancestral graph is *maximal* (MAG) if for any two non-adjacent vertices there is a set that separates them. A *directed acyclic graph* (DAG) is a special kind of MAG, containing only $\rightarrow$ edges, for which m-separation reduces to the familiar d-separation criterion. The *Markov property* links the structure of an ancestral graph $\mathcal{G}$ to observed probabilistic independencies: $X \perp\!\!\!\perp Y \mid \mathbf{Z}$, if $X$ and $Y$ are m-separated by $\mathbf{Z}$. *Faithfulness* implies that the only observed independencies in a system are those entailed by the Markov property. For more details, see (Koller and Friedman, 2009; Spirtes et al., 2000).

An important concept is that of a *minimal* conditional (in)dependence, capturing the notion that really all variables in the minimal set, indicated by brackets, play a role in making two variables (in)dependent:

- $X \perp\!\!\!\perp Y \mid \mathbf{W} \cup [\mathbf{Z}] \equiv \forall \mathbf{Z}' \subsetneq \mathbf{Z} \colon X \not\perp\!\!\!\perp Y \mid \mathbf{W} \cup \mathbf{Z}'$,
- $X \not\perp\!\!\!\perp Y \mid \mathbf{W} \cup [\mathbf{Z}] \equiv \forall \mathbf{Z}' \subsetneq \mathbf{Z} \colon X \perp\!\!\!\perp Y \mid \mathbf{W} \cup \mathbf{Z}'$.

Finally, three path definitions that appear in the orientation rules in the next section: in a MAG, a path $u = \langle X, \ldots, W, Z, Y \rangle$ is a *discriminating path* for $Z$ if $X$ is not adjacent to $Y$, and every node between $X$ and $Y$ is a collider along $u$, and is a parent of $Y$. In a PAG $\mathcal{P}$ (see below), a path $u = \langle V_0, \ldots, V_{n+1} \rangle$ is said to be an *uncovered potentially directed (p.d.) path*, if each successive triple along $u$ is unshielded, and no edge $V_i * - * V_{i+1}$ has an arrowhead at $V_i$ or a tail at $V_{i+1}$. If all edges on $u$ are of the form $\circ\!\!-\!\!\circ$, then the path is called an *uncovered circle path*.

### 2.2 Causal models and ancestral graphs

A popular and intuitive way of representing a causal system is in the form of a causal DAG $\mathcal{G}_C$, where the arrows represent direct causal interactions between variables in a system (Pearl, 2000; Zhang, 2008b). We say there is a causal relation $X \Rightarrow Y$, iff there is a directed path from $X$ to $Y$ in $\mathcal{G}_C$. Absence of such a path is denoted $X \not\Rightarrow Y$. The following properties follow readily from this definition

**Proposition 1.** Causal relations in a DAG $\mathcal{G}_C$ are:

$$\begin{array}{lll} \textit{irreflexive} & : X \Rightarrow X & \vdash \textit{false} \\ \textit{acyclic} & : X \Rightarrow Y & \vdash Y \not\Rightarrow X \\ \textit{transitive} & : (X \Rightarrow Y) \land (Y \Rightarrow Z) \vdash X \Rightarrow Z \end{array}$$

Other definitions are possible, for example we may want to allow for non-recursive relations (feedback) or include threshold effects. Such extensions imply that the causal system is not faithful to a causal DAG, which may impact the conclusions in this article.

When some variables in the causal DAG are hidden, or when there is possible selection bias (Spirtes et al., 1999), the independence relations between the observed variables can be represented in the form of a maximal ancestral graph (Richardson and Spirtes, 2002). Roughly speaking, hidden common causes become bi-directed edges, and selection bias on common effects gives undirected edges. The (complete) partial ancestral graph (PAG) represents all invariant features that characterize the equivalence class $[\mathcal{G}]$ of such a MAG, with a tail '−' or arrowhead '>' end mark on an edge, iff it is invariant in $[\mathcal{G}]$, otherwise it has a circle mark '∘', see (Zhang, 2008a). Tails in a PAG are associated with identifiable (define?) direct causal relations, and arrowheads with the absence thereof, (Zhang, 2008b). Figure 1 illustrates the relation between these three types of graphs. Note that when selection bias may be present, an invariant arc in a PAG $\mathcal{P}$ by itself not necessarily implies a causal relation, e.g. link $B \longrightarrow F$ in Figure 1.3). However, if the tail node also has an incoming invariant arrowhead from another node, as for arc $E \longrightarrow F$, then it *does* represent a definite, identifiable causal relation.

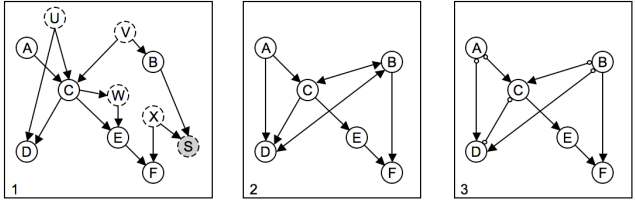

Figure 1: 1) Causal DAG (dashed = hidden, gray = selection); 2) MAG over observed nodes; 3) complete PAG.

The challenge of causal discovery from observed independencies is how to identify all these invariant features from a given data set, in order to determine which variables do or do not have a directed path to which others in the underlying causal DAG.

## 2.3 Augmented FCI algorithm

The famous Fast Causal Inference (FCI) algorithm (Spirtes et al., 2000) was one of the first algorithms that was able to validly infer causal relations from conditional independence statements in the large sample limit, even in the presence of latent and selection variables. It consists of an efficient search for a conditional independence between each pair of variables to identify the skeleton of the underlying causal MAG, followed by an orientation stage to identify invariant tail and arrowhead marks. It was shown to be sound in the large sample limit (Spirtes et al., 1999), although not yet complete. Ali et al. (2005) proved that the seven graphical orientation rules employed by FCI were sufficient to identify all invariant arrowheads in the equivalence class $[\mathcal{G}]$, given a single MAG $\mathcal{G}$. Later, Zhang (2008a) introduced another set of seven rules to orient all remaining invariant tails. Augmented with this set of rules the FCI algorithm is also provably complete.

Loosely speaking, the augmented FCI algorithm consists of an ingenious adjacency search based on conditional independencies (details of which will not concern us here), to find the skeleton of the PAG $\mathcal{P}$, followed by an orientation phase based on a set of graphical rules, detailed in Table 1[1]. Inspection reveals a certain hierarchy in which rules can trigger which others, reflected in the structure of Algorithm 1.

$\mathcal{R}$0a   If $X \perp\!\!\!\perp Y \mid \mathbf{Z}$, then $X \not\!\!\!- Y$, $Sep(X,Y) \leftarrow \mathbf{Z}$.
$\mathcal{R}$0b   If $X *\!\!-\!\!* Z \circ\!\!-\!\!* Y$ and $X \not\!\!\!- Y$, then if $Z \notin Sep(X,Y)$, then $X *\!\!\rightarrow Z \leftarrow\!\!* Y$.
$\mathcal{R}$1    If $X *\!\!\rightarrow Z \circ\!\!-\!\!* Y$, and $X \not\!\!\!- Y$, then $Z \longrightarrow Y$.
$\mathcal{R}$2a   If $Z \longrightarrow X *\!\!\rightarrow Y$ and $Z *\!\!-\!\!\circ Y$, then $Z *\!\!\rightarrow Y$.
$\mathcal{R}$2b   If $Z *\!\!\rightarrow X \longrightarrow Y$ and $Z *\!\!-\!\!\circ Y$, then $Z *\!\!\rightarrow Y$.
$\mathcal{R}$3    If $X *\!\!\rightarrow Z \leftarrow\!\!* Y$, $X *\!\!-\!\!\circ W \circ\!\!-\!\!* Y$, $X \not\!\!\!- Y$, and $W *\!\!-\!\!\circ Z$, then $W *\!\!\rightarrow Z$.
$\mathcal{R}$4a   If $u = \langle X, .., Z_k, Z, Y \rangle$ is a discriminating path between $X$ and $Y$ for $Z$, and $Z \circ\!\!-\!\!* Y$, then if $Z \in Sep(X,Y)$, then $Z \longrightarrow Y$.
$\mathcal{R}$4b   Idem, if $Z \notin Sep(X,Y)$ then $Z_k \leftrightarrow Z \leftrightarrow Y$.
$\mathcal{R}$5    If $u = \langle Z, X, .., W, Y, Z, X \rangle$ is an uncov. circle path, then $Z \text{---} Y$ (idem for all edges on $u$).
$\mathcal{R}$6    If $X \text{---} Z \circ\!\!-\!\!* Y$, then orient as $Z \text{---}* Y$.
$\mathcal{R}$7    If $X \text{---}\!\circ Z \circ\!\!-\!\!* Y$, and $X \not\!\!\!- Y$, then $Z \text{---}* Y$.
$\mathcal{R}$8a   If $Z \longrightarrow X \longrightarrow Y$ and $Z \circ\!\!\rightarrow Y$, then $Z \longrightarrow Y$.
$\mathcal{R}$8b   If $Z \text{---}\!\circ X \longrightarrow Y$ and $Z \circ\!\!\rightarrow Y$, then $Z \longrightarrow Y$.
$\mathcal{R}$9    If $Z \circ\!\!\rightarrow Y$, $u = \langle Z, X, W, .., Y \rangle$ is an uncov. p.d. path, and $X \not\!\!\!- Y$, then $Z \longrightarrow Y$.
$\mathcal{R}$10   If $Z \circ\!\!\rightarrow Y$, $X \longrightarrow Y \longleftarrow W$, $u_1 = \langle Z, S, .., X \rangle$ and $u_2 = \langle Z, V, .., W \rangle$ are uncov. p.d. paths, (possibly with $S = X$ and/or $V = W$), then if $S \not\!\!\!- V$, then $Z \longrightarrow Y$.

Table 1: Orientation rules of augmented FCI

Starting from the fully $\circ\!\!-\!\!\circ$ connected graph in line 1, $\mathcal{R}$0a eliminates all edges between conditionally independent nodes to obtain the skeleton of $\mathcal{P}$ with only $\circ\!\!-\!\!\circ$ edges (line 4). Then rules $\mathcal{R}$0b-$\mathcal{R}$4b obtain all invariant arrowheads (as well as some tails). Rules $\mathcal{R}$5 − $\mathcal{R}$10 then suffice to identify all and only the remaining invariant tails. For example, in Figure 1, the arrowheads at $C$ from $A$ and $B$ are identified by $\mathcal{R}$0b, and the tailmark at $B \longrightarrow F$ follows from $\mathcal{R}$9.

---
[1]We follow the numbering from (Zhang, 2008a). The metasymbol ∗ stands for an arbitrary edge mark, and $X \not\!\!\!- Y$ explicitly indicates the absence of an edge between $X$ and $Y$ in the PAG $\mathcal{P}$; see also Figures 2 and 3.

```
    Input   : independence oracle for V
    Output  : complete PAG P over V
 1: P ← fully ∘−∘ connected graph over V
 2: for all {X, Y} ∈ V do
 3:     search in some clever way for a X ⊥⊥ Y | Z
 4:         P ← R0a  (eliminate X ⊀⊁ Y)
 5:         record Sep(X, Y) ← Z
 6: end for
 7: P ← R0b  (unshielded colliders)
 8: repeat P ← R1 − R4b until finished
 9: P ← R5  (uncovered circle paths)
10: repeat P ← R6 − R7 until finished
11: repeat P ← R8a−R10 until finished
```

**Algorithm 1:** Augmented FCI algorithm

## 3 Inference from causal logic

Note: we use $X$, $Y$, $\mathbf{Z}$, etc. to denote disjoint (sets of) observed variables, and $\mathbf{S}$ to denote the (possibly empty) set of selection nodes in a causal DAG $\mathcal{G}_C$.

### 3.1 Logical rules from minimal independence

There is a well-known, fundamental connection between minimal (in)dependencies and causal relations:

**Lemma 2.** If a node $Z$ changes an (in)dependence relation between $X$ and $Y$ in a causal DAG, then:

1. $X \perp\!\!\!\perp Y \,|\, \mathbf{W} \cup [Z] \;\;\vdash\;\; Z \Rightarrow (X \cup Y \cup \mathbf{W} \cup \mathbf{S})$,

2. $X \not\perp\!\!\!\perp Y \,|\, \mathbf{W} \cup [Z] \;\;\vdash\;\; Z \not\Rightarrow (X \cup Y \cup \mathbf{W} \cup \mathbf{S})$.

with special case $X \perp\!\!\!\perp Y \,|\, [\mathbf{W} \cup Z] \vdash Z \Rightarrow (X \cup Y \cup \mathbf{S})$.

This means that, using $X \not\Rightarrow Y \stackrel{\text{def}}{=} \neg(X \Rightarrow Y)$, we can translate observed minimal (in)dependencies *directly* into logical statements about causal relations:

**Lemma 3.** For observed minimal (in)dependencies between nodes in a causal DAG $\mathcal{G}_C$:

1. $X \perp\!\!\!\perp Y \,|\, [\mathbf{W} \cup Z] \;\;\vdash\;\; Z \Rightarrow X \;\vee\; Z \Rightarrow Y \;\vee\; Z \Rightarrow \mathbf{S}$

2. $X \not\perp\!\!\!\perp Y \,|\, \mathbf{W} \cup [Z] \;\;\vdash\;\; Z \not\Rightarrow X \;\wedge\; Z \not\Rightarrow Y \;\wedge\;$
$\phantom{X \not\perp\!\!\!\perp Y \,|\, \mathbf{W} \cup [Z] \;\;\vdash\;\;} Z \not\Rightarrow \mathbf{W} \;\wedge\; Z \not\Rightarrow \mathbf{S}$

By establishing which minimal (in)dependencies hold in a distribution, a list L can be compiled of logical statements of the form:

1: $Z \Rightarrow X \;\vee\; Z \Rightarrow Y \;\vee\; Z \Rightarrow \mathbf{S}$
2: $X \not\Rightarrow Y$
3: $Y \Rightarrow X \;\vee\; Y \Rightarrow W \;\vee\; Y \Rightarrow \mathbf{S}$, etc.

Each line states a truth, for one specific node, about the causal relations it has with one or more others. New statements can be inferred by substituting the subject of one line in another, and then reduce by using the three causal properties from Proposition 1.

To illustrate the inference process in deriving (new) causal information, consider these two examples:

**Example 1.** Suppose in a causal system $\mathcal{G}_C$ both $X \perp\!\!\!\perp Y \,|\, [\mathbf{Z}]$ and $X \not\perp\!\!\!\perp U \,|\, \mathbf{W} \cup [Z]$ have been observed, for some $Z \in \mathbf{Z}$. Then this corresponds to

1: $Z \Rightarrow X \;\vee\; Z \Rightarrow Y \;\vee\; Z \Rightarrow \mathbf{S}$
2: $Z \not\Rightarrow X \;\wedge\; Z \not\Rightarrow U \;\wedge\; Z \not\Rightarrow \mathbf{S} \;\wedge\; Z \not\Rightarrow \mathbf{W}$

Using (2:) to eliminate $Z \Rightarrow X$ and $Z \Rightarrow \mathbf{S}$ from (1:) then gives (3:)

$\vdash \;\; (false) \;\vee\; Z \Rightarrow Y \;\vee\; (false)$
3: $\phantom{\vdash \;\; (false) \;\vee\;} Z \Rightarrow Y$

This case corresponds to the embedded Y-structure from Mani et al. (2006), and matches the conditions for orientation rule $\mathcal{R}1$.

**Example 2.** Suppose in a causal system $\mathcal{G}_C$ both $Z \perp\!\!\!\perp W \,|\, [\mathbf{U}_{ZW} \cup X]$ and $X \perp\!\!\!\perp Y \,|\, [\mathbf{U}_{XY} \cup Z \cup W]$ have been observed, with $\mathbf{U}_{XY}$ and $\mathbf{U}_{ZW}$ two possibly empty/overlapping sets of nodes. Then for the inference list this gives statement (1:) from the first independence, and (2:) and (3:) from the second:

1: $X \Rightarrow Z \;\vee\; X \Rightarrow W \;\vee\; \phantom{Z \Rightarrow Y \;\vee\;} X \Rightarrow \mathbf{S}$
2: $Z \Rightarrow X \;\vee\; \phantom{Z \Rightarrow W \;\vee\;} Z \Rightarrow Y \;\vee\; Z \Rightarrow \mathbf{S}$
3: $\phantom{Z \Rightarrow X \;\vee\;} W \Rightarrow X \;\vee\; W \Rightarrow Y \;\vee\; W \Rightarrow \mathbf{S}$

Using transitivity and irreflexivity, when substituting (2:) and (3:) in (1:), this reduces to (4:)

$\vdash \;\; X \Rightarrow X \;\vee\; X \Rightarrow X \;\vee\; X \Rightarrow Y \;\vee\; X \Rightarrow \mathbf{S}$
4: $\phantom{\vdash \;\; X \Rightarrow X \;\vee\; X \Rightarrow X \;\vee\;} X \Rightarrow Y \;\vee\; X \Rightarrow \mathbf{S}$

This case matches instances of $\mathcal{R}9$, where all alternatives for $X$ from the first minimal independence necessarily lead to a causal relation to node $Y$ (or $\mathbf{S}$). However, contrary to Example 1, selection bias cannot be eliminated from these two statements alone.

### 3.2 Inferred statements

Remarkably enough, Lemma 3 and Proposition 1 are already sufficient to identify almost all causal information that can be discovered from probabilistic independencies, by repeatedly executing the substitute and reduce steps on the list of logical statements L. There is just one more piece of information needed to complete the puzzle.

**Lemma 4 (Inferred blocking node).** In a causal system $\mathcal{G}_C$, if $X \perp\!\!\!\perp Y \,|\, [\mathbf{Z}]$, and there is a subset $\{Z_1, \ldots, Z_k, Z\} \subseteq \mathbf{Z}$, such that in the sequence $[\mathbf{U}] \equiv [U_0, \ldots, U_{k+2}] = [X, Z_1, \ldots, Z_k, Z, Y]$ it holds that:

- $U_i \not\Rightarrow \{U_{i-1}, U_{i+1}\}$,
- $U_j \not\perp\!\!\!\perp U_{j+1} \,|\, \mathbf{Z}'$,

with $i = 1..k$, and with $j = 0..(k + 1)$ and $\forall \mathbf{Z}' \subseteq \mathbf{Z} \setminus \{U_j, U_{j+1}\}$, then $Z \Rightarrow (Z_k \cup Y \cup \mathbf{S})$.

In other words, if we find an 'inferred blocking node', then we can add the following statement to the list:

1: $\quad Z \Rightarrow Z_k \;\vee\; Z \Rightarrow Y \;\vee\; Z \Rightarrow \mathbf{S}$

Lemma 4 is clearly a generalization of rule $\mathcal{R}$4a: if the nodes in the sequence $[\mathbf{U}]$ are adjacent in $\mathcal{P}$, then it corresponds to a discriminating path for $Z$, and the non-independence tests in the second item can be omitted. Note that resulting statement (1:) reveals that the discriminating path for $Z$ in $\mathcal{R}$4a behaves identical to a node $Z$ observed in a minimal independence between $Z_k$ and $Y$. As a result, whether or not we observe $Z_k \perp\!\!\!\perp Y \,|\, [.. \cup Z]$, the fact that in the given conditions $Z$ does *not* create a dependency between $X$ and $Y$, allows us to infer that $Z$ blocks some path between $Z_k$ and $Y$; hence 'inferred blocking node'.

One remark: the set of possible independence relations involved in lemma 4 may seem quite daunting. However, in section 5 we will see that ultimately only a handful need to be checked.

### 3.3 Direct and indirect causal relations

Reasoning with presence or absence of causal relations implies that we are not limited to direct causal influences only, but can draw on other, indirect sources of causal information as well: both can be used to derive new information in exactly the same way. But sometimes it can be very useful to distinguish between direct and indirect causes. In the PAG, a missing edge represents absence of a direct cause.

**Lemma 5.** In a causal system $\mathcal{G}_C$, a (minimal) conditional independence $X \perp\!\!\!\perp Y \,|\, \mathbf{Z}$ implies that all causal paths $X \Rightarrow Y$ or $X \Leftarrow Y$, or common causes of $X$ and $Y$ in $\mathcal{G}_C$ are mediated by nodes in $\mathbf{Z}$.

For independent nodes $X \perp\!\!\!\perp Y \,|\, \varnothing$ it implies neither is a cause of the other: $(X \not\Rightarrow Y) \wedge (Y \not\Rightarrow X)$.

Lemma 5 gives the global structure (skeleton) of the PAG. If we want to distinguish between direct and indirect causal relations, we can simply use the MAG definition for tail/arrowhead marks (see Richardson and Spirtes, 2002, §4.2) to project the causal information in the list L onto this skeleton:

**Lemma 6.** The causal information from statements in the list L can be transferred to invariant edge marks between adjacent nodes in the corresponding PAG $\mathcal{P}$:

  - if $X \not\Rightarrow Y \in$ L, then $X \leftarrow\!\!* \, Y$,
  - if $X \Rightarrow Y \;(\vee\; X \Rightarrow \mathbf{S}) \in$ L, then $X \,\text{---}\!* \, Y$,

In the next section we will see that this is also complete. It means that after the logical causal inference (LoCI) process has completed, we can optionally *choose* to reproduce the PAG, provided the global structure could be established. If not, for example because only an arbitrary subset of (in)dependence relations was available, then all causal information in the list L remains valid, even though the orientation rules in Table 1 can no longer be applied.

## 4 A logical characterization of causal information

In this section we show that the combination of the logical statements, derived directly from observed/inferred conditional (in)dependencies via lemmas 3 and 4, together with the three inference rules in proposition 1, are sufficient to obtain all invariant orientations (tails and arrowheads) in the PAG. We do this by matching each orientation rule to specific instances of the lemmas, and already inferred information. In doing so, we make good use of the known completeness of the augmented FCI algorithm.

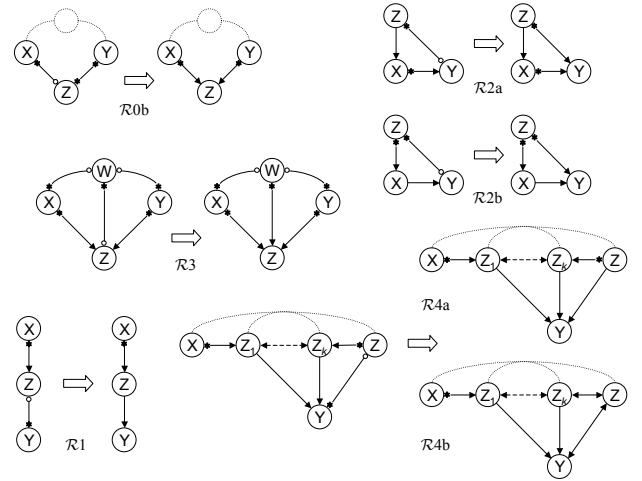

Figure 2: Rules $\mathcal{R}$0b–$\mathcal{R}$4b, arrowhead orientation rules

### 4.1 Invariant arrowheads

First we show that all graphical orientation rules that can identify invariant arrowheads, see Figure 2, are, in fact, different graphical instances of just two cases, that can be found from minimal independencies and subsequent dependencies.

We would like to emphasize that there is no need to search for the specific cases discussed in this section: they automatically pop up when running the causal logic rules in proposition 1 on the list of statements L. We use them here to characterize all causal information that can be identified in this way, and thus, since the augmented FCI algorithm is complete, by any algorithm for causal discovery.

**Lemma 7.** In a PAG $\mathcal{P}$, all invariant arrowheads $Z \ast\!\!\to Y$ are instances of

(1): $U \not\!\perp\!\!\!\perp V \,|\, \mathbf{W} \cup [Y]$, created from $U \perp\!\!\!\perp V \,|\, [\mathbf{W}]$, with $Z \in (U \cup V \cup \mathbf{W})$,

(2): $X \perp\!\!\!\perp Y \,|\, [\mathbf{W} \cup Z]$, with $Z \not\Rightarrow (X \cup \mathbf{S})$ from either case (1) or case (2).

In words: all invariant arrowheads originate from either an observed conditional dependence (1), or as the reverse of a definite causal relation (2).

As a result, all seven arrowhead orientation rules $\mathcal{R}0-\mathcal{R}4b$ are covered by lemma 3. Note that, when starting from the full set of (in)dependence statements in lemma 3, it is *not* necessary to consider discriminating paths (nor 'inferred blocking nodes'), in order to guarantee arrowhead completeness, contrary to when starting from a MAG, as in (Ali et al., 2005).[2]

### 4.2 Invariant tails

The previous section will not only find all arrowheads, but also a number of invariant tails, as case (2) in lemma 7 already covers all instances of rule $\mathcal{R}1$, including the tail $Z \longrightarrow Y$. In this section we show that all remaining invariant tails from rules $\mathcal{R}5 - \mathcal{R}10$, see Figure 3, correspond to three cases, that can be found from minimal independencies in combination with the three inference rules in proposition 1.

To do that, we first introduce the following concept:

**Definition.** A *transitive relation* from $X$ to $Y$ is a sequence of nodes $[X, Z_1, \ldots, Z_k, V_1, \ldots, V_m, Y]$ (not necessarily distinct), such that:

- $\forall Z_i, \exists \mathbf{U}_i : Z_{i-1} \perp\!\!\!\perp Z_{i+1} \,|\, [\mathbf{U}_i \cup Z_i]$,
- $\forall V_j : V_j \Rightarrow (V_{j+1} \cup \mathbf{S})$,

with $Z_0 = X, Z_{k+1} = V_1, V_{m+1} = Y$, for $k, m \geq 0$.

In words: a series of overlapping minimal conditional independencies, followed by a causal relation. A transitive relation can be as short as a single independence $X \perp\!\!\!\perp Y \,|\, [Z_1]$, or a relation $X \Rightarrow Y$. As such, it is a generalization of the uncov. p.d. path in section 2.3.

The reason for this introduction is the property:

**Corollary 8.** In a causal system $\mathcal{G}_C$, if there is a transitive relation $[X, Z_1, \ldots, Y]$, then:

- $X \Rightarrow (Z_1 \cup \mathbf{S}) \;\vdash\; X \Rightarrow (Y \cup \mathbf{S})$.

We can now state:

**Lemma 9.** In a PAG $\mathcal{P}$, all invariant tails $Z \longrightarrow\ast Y$ from graphical orientation rules $\mathcal{R}4a, \mathcal{R}5, \mathcal{R}7, \mathcal{R}9$, and $\mathcal{R}10$ are instances of:

(2b): $X \perp\!\!\!\perp Y \,|\, [\mathbf{W} \cup Z]$, with $X \Rightarrow (Z \cup \mathbf{S})$ from either case (3) or another instance of (2b),

(3): $U \perp\!\!\!\perp V \,|\, [\mathbf{W} \cup W]$, with two transitive relations $[W, U, .., Y] + [W, V, .., Y]$, and $Z \in \{U, V, W\}$,

(4): $X \perp\!\!\!\perp Y \,|\, [\mathbf{Z}]$, with inferred blocking node $Z \in \mathbf{Z}$, together with $Z_k \Rightarrow (Y \cup \mathbf{S})$ from either case (2) or case (4).

Case (2b) covers rule $\mathcal{R}7$, and is so named because of its similarity/overlap with case (2) for $\mathcal{R}1$. Case (3) covers all instances of rules $\mathcal{R}5, \mathcal{R}9$, and $\mathcal{R}10$, and case (4) accounts for tails from orientation rule $\mathcal{R}4a$. In most instances of case (3) the transitive relation requires only a single minimal conditional independence, even for long paths. Often, both transitive relations can be captured together in a single independence, as in Example 2.

A nice property is that all identifiable selection nodes $X \Rightarrow \mathbf{S}$ also pop out 'automatically' by applying the inference rules in lemma 1 on instances of case (3):

**Corollary 10.** In a PAG $\mathcal{P}$, all identifiable selection nodes $X \Rightarrow \mathbf{S}$ are covered by case (3), in the form of a minimal independence with two transitive relations back to itself.

That leaves just tails from three more orientation rules to handle. However, these too follow implicitly from the existing cases:

**Corollary 11.** In a PAG $\mathcal{P}$, all invariant tails from orientation rules $\mathcal{R}6, \mathcal{R}8a$, and $\mathcal{R}8b$, are covered by the causal logic rules applied to cases (1)-(4).

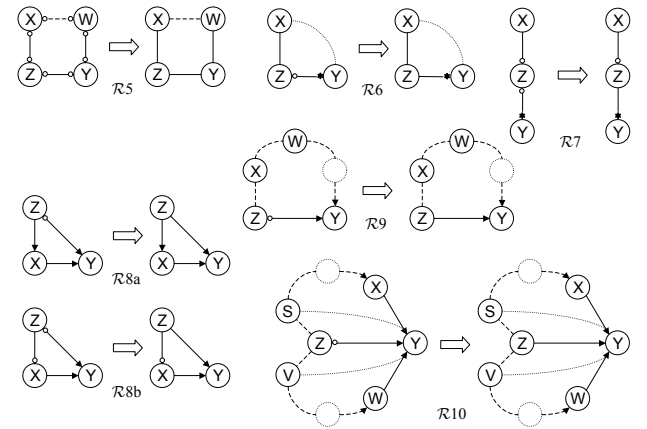

Figure 3: Rules $\mathcal{R}5 - \mathcal{R}10$, tail orientation rules

---

[2]This may seem contradictory, as a MAG is just an encoding of an independence model, but it is not possible to read *which* set separates $X$ and $Y$ in $\mathcal{R}4a/b$ from the MAG, without actually checking for the discriminating path.

# 5 Reconstructing the PAG

In this section we look at the logical inference process itself, and provide an efficient anytime algorithm for deriving the PAG.

## 5.1 Inference procedure

A nice property is that the logical substitute/reduce steps take on a particularly simple form: it only involves statements that are a logical disjunction of at most two causal relations and possible selection bias, or a single term for the absence of a causal relation. In other words, the list Ł always keeps the form in section 3.1. Each step consists of a substitution of one statement in another followed by a reduction to this standard form. Furthermore, as more information becomes available, statements in the list can simplify from three to two or even one term. Cf. example 1, where inferred statement (3:) replaces (1:), as there is no point in keeping the original.

The next result limits the independence search:

**Lemma 12.** In the logical causal inference (LoCI) approach, finding a *single*, arbitrary $X \perp\!\!\!\perp Y \mid [\mathbf{Z}]$, for each pair of nodes $(X, Y)$ in the graph (if it exists) is sufficient to find all invariant features of the PAG.

Fortunately, the current implementation of the FCI algorithm already finds only *minimal* conditional independencies for each pair of nodes (if it exists), as it looks for sets of increasing size until it finds one that separates the two. (This is also the dominant factor in the time-complexity of the algorithm.) For each pair found, we still need to check for other nodes that can destroy this independence (lemma 3, item 2), however, this is negligible compared to the search itself.

Furthermore, the inferred blocking node from lemma 4, can be tackled efficiently, *after* all invariant arrowheads have been found from cases (1) and (2): it makes it possible to establish the 'non-ancestor' conditions in the sequence in one go. Together with a restriction to a sequence of non-separated nodes (avoiding the additional dependence tests), this greatly reduces the number of candidates to check.

The final step is to use lemmas 5 and 6 to transfer the logical information in the list Ł to invariant edge marks in the skeleton of $\mathcal{P}$.

## 5.2 The LoCI algorithm

We can now give the outline of an algorithm that is able to infer the complete PAG, using the logical causal inference approach described in section 3.

Algorithm 2 borrows the initial search for (minimal)

```
Input   : independence oracle for V
Output  : complete PAG P over V
1: for all {X, Y} ∈ V do
2:     search in some clever way for a X ⊥⊥ Y | [Z]
3:        ∀Z ∈ Z : Ł ← Z ⇒ (X ∪ Y ∪ S)
4:        ∀W, X ⊥̸⊥ Y | Z ∪ W :
5:           Ł ← W ⇏ (X ∪ Y ∪ Z ∪ S)
6:     repeat Ł ← substitute/reduce until finished
7: end for
8: Ł ← Z ⇒ (Z_k ∪ Y ∪ S), ∀Z : inferred block. node
9: repeat Ł ← substitute/reduce until finished
10: P ← fully ∘–∘ connected graph over V
11:    eliminate X ⨯ Y, iff X ⊥⊥ Y | [∗]
12:    orient X —∗ Y, iff X ⇒ (Y ∪ S) ∈ Ł
13:    orient X ←∗ Y, iff X ⇏ Y ∈ Ł
```

**Algorithm 2:** Logical Causal Inference (LoCI) algorithm

conditional independencies from the standard FCI algorithm. If it finds one it is recorded in the list Ł, line 3, and checked for nodes that destroy this independence (also recorded in Ł). Each time a minimal independence has been found, line 6 runs the inference rules to update the identifiable causal information. This step could be run just once, after the independence search has completed, but in practice the impact on performance is negligble and far outweighed by the fact that most causal information is already identifiable (available) in the early stages of the process. At line 8, all non-ancestor relations ($X \not\Rightarrow Y$) have been found (see lemma 7), which makes it relatively easy to find the remaining 'inferred blocking nodes' from lemma 4 in line 8. If any are found that contain new information, then line 9 infers the remaining relations. Finally, lines $10 - 13$ construct the equivalent PAG representation from the list Ł.

These results can now be summarized as:

**Theorem 1.** The Logical Causal Inference (LoCI) algorithm is sound and complete.

# 6 Discussion and conclusion

In this paper we developed a new approach to constraint-based causal discovery: observed minimal (in)dependencies are converted into logical statements about causal relations, and these statements are subsequently combined using basic properties of causality.

It leads to a remarkably simple characterization, in which all identifiable causal relations take the form of an (inferred) minimal conditional independence with either elimination of one alternative, or both alternatives leading to the same conclusion.

The resulting logical causal inference (LoCI) method was put to work in an efficient anytime algorithm, the first alternative to the augmented FCI-algorithm shown to be both sound and complete. The LoCI algorithm is strikingly simpler than its counterpart in section 2.3. Even though it is not necessarily faster, as for both the overall complexity is dominated by the independence search, the fact that the implementation takes on this very simple and elegant form suggests it is somehow more 'natural' to causal discovery.

The way in which the LoCI algorithm builds up this causal information is markedly different from many other constraint-based methods: instead of focussing on combinations of node-pairs that may or may not be separable (the essence of graphical orientation rules), the LoCI algorithm focusses on the nodes that separate them. In particular, as it does not need to search for pairs of nodes that cannot be separated by any set (the edges forming the skeleton of the PAG), the approach taken by the algorithm could be dubbed 'structure independent'. As a result, it can be adapted to search for target causal relations in large models, updating each time as new independence information becomes available; of course, if we want to ensure completeness, we still have to find all of them.

The simplicity of the LoCI algorithm raises the question if a similar approach is viable in other applications as well. For example, incorporation of causal information from background knowledge or derived from other properties of the distribution (Shimizu et al., 2006; Mooij et al., 2010), is straightforward. The same holds for additional assumptions, such as 'no selection bias'. Including interventional information also fits nicely in this framework, and requires only minor modifications of the minimal independence lemma 2. An extension to multiple models, similar to (Triantafillou et al., 2010; Claassen and Heskes, 2010), seems feasible as well. To prove completeness, we had to rely on the known completeness of the augmented FCI algorithm, but we suspect that a more direct proof should be possible.

Perhaps the most promising aspect of the LoCI approach lies in the flexibility it offers in deriving causal information. For example, we are free to ignore any suspect, 'borderline' (in)dependence decisions, by not including them in the list L in lines 3 and 5: all inferred causal relations remain valid. This should definitely increase the reliability of the output, even though it is no longer guaranteed to be complete. Finally, the 'structure independent' aspect implies there are many different ways to arrive at the same conclusion. This makes it possible to choose the most reliable combination(s) of independencies for a more robust conclusion and to detect inconsistencies. Tracking which logical statements in L are combined to identify new relations could also improve accountability for the output, indicating exactly *why* a causal relation was found.

### Acknowledgement

This research was supported by VICI grant 639.023.604 from the Netherlands Organization for Scientific Research (NWO).

## Appendix A. Proofs

This section contains the key steps of all proofs; for details, see supplement (Claassen and Heskes, 2011).

**Lemma 2.**
*Proof sketch.* A variant of two well-known results, see (Spirtes et al., 1999; Claassen and Heskes, 2010).
(1.) If $Z$ blocks the (final) unblocked path between $X$ and $Y$ given $\mathbf{W}$, then it must be a noncollider on a trek between two of the other nodes involved; hence a directed path in $\mathcal{G}_C$, and so a causal relation.
(2.) A node can only unblock a path, if there are unblocked paths into that node given the others. Therefore, if $Z$ has a directed path to any $(\mathbf{W} \cup \mathbf{S})$, then conditioning on $Z$ is not needed to unblock the path, and if it has a directed path to $X$ or $Y$ then without $Z$ there is already an unblocked path (via $Z$).
The special case follows from (1.) and acyclicity. □

**Lemma 4**
*Proof.* In words: if no node $Z_i$ has a causal relation (directed path in $\mathcal{G}_C$) to either of its neighbors in the sequence $[X, Z_1, \ldots, Z_k, Z, Y]$, and all neighboring nodes in the sequence are dependent given any subset of $\mathbf{Z}$, then $Z$ has a causal relation to $Z_k$, $Y$, and/or $\mathbf{S}$. By construction, in $\mathcal{G}_C$ there is an unblocked path from $X$ to $Z$, given $\mathbf{Z}$. If both $Z_k$ and $Y$ have paths that are into $Z$, then the sequence would represent an unblocked path between $X$ and $Y$ given $\mathbf{Z} \cup Z$ in $\mathcal{G}_C$, which would make $X$ and $Y$ dependent, contrary the given. By the second item (dependent neigbors), all neighbors in the sequence, so also $(Z_k, Z)$ and $(Z, Y)$, are connected by treks between them (or treks to $\mathbf{S}$), that are not blocked by any nodes from $\mathbf{Z}$. Not both these paths from $Z_k$ and $Y$ are into $Z$, therefore $Z$ must either have a directed path to $\mathbf{S}$ in $\mathcal{G}_C$ and/or be an ancestor of $Z_k$ or $Y$. □

**Lemma 7**
*Proof sketch.* Both cases are sound:
(1.) By lemma 3, item 2, the first gives $(Y \not\Rightarrow Z) \land (Y \not\Rightarrow \mathbf{S})$, which, by definition, implies that if $Y$ has an edge to $Z$ in $\mathcal{P}$, then the mark at $Y$ is an (invariant) arrowhead.
(2.) The second is an application of lemma 3, item

1, giving $(Z \Rightarrow X) \vee (Z \Rightarrow Y) \vee (Z \Rightarrow \mathbf{S})$, where the first and third are eliminated by the arrowhead at $X \ast\!\!\rightarrow Z$ (def). Therefore $Z \Rightarrow Y$, and so (acyclicity) also $Y \not\Rightarrow Z$, but also $Y \not\Rightarrow \mathbf{S}$, otherwise (transitivity) $Z \Rightarrow \mathbf{S}$. Therefore, if $Y$ has an edge to $Z$ in $\mathcal{P}$, then it has an arrowhead mark at $Y$.

The proof that they are also complete follows by induction on the graphical orientation rules $\mathcal{R}$0b–$\mathcal{R}$4b, showing that none of them introduces a violation of lemma 7. As these rules are sufficient for arrowhead completeness, it follows that the lemma holds for all invariant arrowheads. □

**Corollary 8**
*Proof.* The transitive relation implies:
$$\begin{array}{ll}1: & Z_1 \Rightarrow X \quad \vee \; Z_1 \Rightarrow Z_2 \; \vee \; Z_1 \Rightarrow \mathbf{S} \\ k: & Z_k \Rightarrow Z_{k-1} \vee \; Z_k \Rightarrow V_1 \; \vee \; Z_k \Rightarrow \mathbf{S} \\ k{+}m: & V_m \Rightarrow Y \quad \vee \; V_m \Rightarrow \mathbf{S}\end{array}$$
Back substituting in reverse order gives finally,
$$\vdash \quad Z_1 \Rightarrow X \quad \vee \; Z_1 \Rightarrow Y \vee Z_1 \Rightarrow \mathbf{S}$$
to substitute in $X \Rightarrow Z_1 \vee X \Rightarrow \mathbf{S}$. □

**Lemma 9**
*Proof sketch.* All three cases are sound:
(2b) By lemma 3, $X \perp\!\!\!\perp Y \,|\, [\mathbf{W} \cup Z]$ gives $(Z \Rightarrow X) \vee (Z \Rightarrow Y) \vee (Z \Rightarrow \mathbf{S})$. Combined with $X \Rightarrow Z \vee X \Rightarrow \mathbf{S}$ this reduces to $(Z \Rightarrow Y) \vee (Z \Rightarrow \mathbf{S})$, and so a tail at $Z$ if it has an edge to $Y$ in $\mathcal{P}$.
(3) Idem, $U \perp\!\!\!\perp V \,|\, [\mathbf{W} \cup W]$ gives $(W \Rightarrow U) \vee (W \Rightarrow V) \vee (W \Rightarrow \mathbf{S})$. From corollary 8, the transitive relations give $(W \Rightarrow \{U \cup \mathbf{S}\}) \vdash (W \Rightarrow \{Y \cup \mathbf{S}\})$, and $(W \Rightarrow \{V \cup \mathbf{S}\}) \vdash (W \Rightarrow \{Y \cup \mathbf{S}\})$. Substituting these two in the first then gives $(W \Rightarrow Y) \vee (W \Rightarrow \mathbf{S})$. This holds for all nodes on the two transitive chains, hence if $Z \in \{U, V, W\}$, then $(Z \Rightarrow Y) \vee (Z \Rightarrow \mathbf{S})$, and therefore a tail $Z \!-\!\ast Y$, if they are connected in $\mathcal{P}$.
(4) By lemma 4, as $Z$ is an inferred blocking node between $X$ and $Y$ given $\mathbf{Z}$, there is a $Z_k \in \mathbf{Z}$ such that $Z \Rightarrow Z_k \vee Z \Rightarrow Y \vee Z \Rightarrow \mathbf{S}$. Together with the given $Z_k \Rightarrow Y \vee Z_k \Rightarrow \mathbf{S}$, this reduces to $Z \Rightarrow Y \vee Z \Rightarrow \mathbf{S}$, and hence an invariant tail $Z \longrightarrow Y$.

For completeness it is fairly straightforward to see that in a PAG $\mathcal{P}$, all instances of rule $\mathcal{R}$7 match (2b), instances of rules $\mathcal{R}$5, $\mathcal{R}$9, and $\mathcal{R}$10 always match case (3), and $\mathcal{R}$4a matches case (4). □

**Corollary 10**
*Proof sketch.* By corollary 8 a transitive relation $[W, U, .., W]$ implies that $W \Rightarrow (U \cup \mathbf{S}) \;\vdash\; W \Rightarrow \mathbf{S}$. Idem for $[W, V, .., W]$. Two such statements connected by $U \perp\!\!\!\perp V \,|\, [\mathbf{W} \cup W]$ then reduce $(W \Rightarrow U) \vee (W \Rightarrow V) \vee (W \Rightarrow \mathbf{S})$, from lemma 3, to $(W \Rightarrow \mathbf{S})$, i.e. identifiable selection bias. That all nodes with identifiable selection bias have this form follows from the fact that only rules $\mathcal{R}5 - \mathcal{R}7$ can produce undirected edges on nodes (corresponding to identifiable selection). The uncovered circle path from $\mathcal{R}$5 already has this form; $\mathcal{R}$6 can be ignored when identifying new nodes, and $\mathcal{R}$7 can only produce undirected edges on transitive chains connecting two distinct circle-path components. □

**Corollary 11**
*Proof.* $\mathcal{R}$8a and $\mathcal{R}$8b do not take (in)dependence information as input, but only need lemma 1 to combine two relations already found as a result of cases (1)-(4). The tail from rule $\mathcal{R}$6 only signifies that $Z \Rightarrow \mathbf{S}$, which will be found as case (3), by corollary 10. □

**Lemma 12**
*Proof sketch.* We know from lemmas 7 and 9 that all orientation rules are covered by *some* combination of minimal independencies and subsequent dependencies. From the graphical description, we see that the rules orient tails/arrowheads between adjacent nodes that either involve a noncollider between two nonadjacent nodes, (and are therefore part of *all* minimal conditional independencies between the two), or as one of the separated nodes in the conditional independence (so any will do), possibly with a direct link to a node that destroys this independence (which will therefore also be found). The only rule that is not entirely straightforward is $\mathcal{R}$2a, but this boils down to a similar case as for the invariant arrowheads in lemma 7, to which the same argument can be applied, to show that the node $Z$ is a necessary part of at least *some* minimal independence that is destroyed by $Y$. Therefore all rules are covered, if we have at least one minimal independence for each pair of nonadjacent nodes, in combination with the subsequent dependencie. As we know that the set of graphical orientation rules is sufficient to find all invariant features in the PAG (Zhang, 2008a), this proves the lemma. □

**Theorem 1**
*Proof sketch.* Soundness follows from the validity of the lemmas 3 and 4, that produce the logical statements in the list Ł, in combination with the causal logic rules in lemma 1. Completeness follows from the fact that all rules are instances of cases (1)-(4) for a single, arbitrary minimal independence between nodes, in combination with subsequent dependencies (lemma 11), the fact that all logical inference in each of the cases (1)-(4) is covered by lemma 3, the fact that case (1) and (2) will find all required non-ancestor relations (= invariant arrowheads, see Zhang, 2008a, lemma 6), needed to obtain the only remaining piece of information (inferred blocking node for case (4) from lemma 4). After running the logical rules on this set of statements to completion, all invariant edge marks have been found and can be transferred to the PAG. □